\documentclass{article} 
\usepackage{iclr2020_conference,times}

\usepackage{amsmath,amsfonts,bm}

\def\eqref#1{equation~\ref{#1}}

\def\1{\bm{1}}

\DeclareMathAlphabet{\mathsfit}{\encodingdefault}{\sfdefault}{m}{sl}
\SetMathAlphabet{\mathsfit}{bold}{\encodingdefault}{\sfdefault}{bx}{n}

\usepackage{hyperref}
\usepackage{url}
\usepackage{graphicx}
\usepackage{tabularx}
\usepackage{booktabs}
\usepackage{enumitem}
\usepackage{subcaption}

\title{ANEA: Distant Supervision for Low-Resource Named Entity Recognition}

\author{\textbf{Michael A. Hedderich}\textsuperscript{1}, \textbf{Lukas Lange}\textsuperscript{2} \& \textbf{Dietrich Klakow}\textsuperscript{1}\\
\textsuperscript{1}Saarland University, Saarland Informatics Campus, Germany\\
\textsuperscript{2}Bosch Center for Artificial Intelligence, Germany\\
\texttt{\{mhedderich,dietrich.klakow\}@lsv.uni-saarland.de} \\
\texttt{lukas.lange@de.bosch.com}
}

\iclrfinalcopy 
\begin{document}

\maketitle
\begin{abstract}
Distant supervision allows obtaining labeled training corpora for low-resource settings where only limited hand-annotated data exists. However, to be used effectively, the distant supervision must be easy to gather. In this work, we present ANEA, a tool to automatically annotate named entities in texts based on entity lists. It spans the whole pipeline from obtaining the lists to analyzing the errors of the distant supervision. A tuning step allows the user to improve the automatic annotation with their linguistic insights without labelling or checking all tokens manually. In six low-resource scenarios, we show that the F1-score can be increased by on average 18 points through distantly supervised data obtained by ANEA.
\end{abstract}

\section{Introduction}

Named Entity Recognition (NER) is a core NLP task necessary for various applications, from information retrieval to virtual assistants. While there exist some large, hand-annotated corpora like \citep{tjong2003conll} or \citep{weischedel2011ontonotes}, these are limited to a selected set of languages and domains. For many low-resource languages and domains, it is not possible to manually label every token of large corpora due to time and resource constraints. The absence of labeled data is prevalent for languages from developing countries. We see this as a significant factor limiting the development of NLP technologies in these regions with respect to the ongoing tendency towards data-driven models. 

To overcome the lack of labeled data, weak or distant supervision methods have become popular, which automatically annotate unlabeled, raw text. Even in low-resource settings, unlabeled text is often available, and research has shown that automatically annotated labels can be a useful training resource in the absence of expensive, high-quality labels. For NER, a widespread approach is to use lists, dictionaries or gazetteers of named entities (e.g. a list of person names or cities). Each word in the corpus is assigned the corresponding named entity label if it appears in this list of entities. Introduced by \citet{mintz2009distant}, this is still a popular technique and used e.g. by  \citet{peng-etal-2019-distantly}, \citet{adelani2020distant} and \citet{lison-etal-2020-weak-supervision}. For an extensive list of recent works using distant supervision for low-resource NER, we refer to the recent survey by \citet{hedderich2020survey}.

While distant supervision performs very well on high-resource languages, it has been shown to be more difficult to leverage in real low-resource settings due to the lack of external information \citep{kann2020POSPoorly}. Additionally, several difficulties arise  when applying it in a practical way, such as obtaining these dictionaries (e.g. a list of city names in Yorùbá) or adapting the matching procedure to the specific language and domain (e.g. deciding for or against lemmatization and, thus, trading off recall and precision). Distant supervision can only be beneficial and save resources if it is easy to use and fast to deploy. 

The ANEA tool we present provides the functionality to actually use distant supervision approaches in practice for many languages and named entity types while minimizing the amount of manual effort and labeling cost. A process is provided to automatically extract entity names from Wikidata, a free and open knowledge base. The information is used to annotate named entities for large amounts of unlabeled text automatically. The tool also supports the user in tuning the automatic annotation process. It enables language experts to efficiently include their knowledge without having to annotate many tokens manually. Both a library and a graphical user interface are provided to assist users of varying technical backgrounds and different use-cases. In an experimental study on six different scenarios, we show that ANEA outperforms two baselines in nearly all cases regarding the quality of the automatic annotation. When used to provide distantly supervised training data for a neural network model, it creates on average a boost of 18 F1 points with less than 30 minutes of manual interaction. The tool, further information and technical documentation and the additional model code and evaluation data is made publicly available online.\footnote{\url{https://github.com/uds-lsv/anea}}
    
\section{Related Work}

A variety of open-source tools exist to annotate text manually. While their focus is on the manual annotation of data, some support the user with certain degrees of automation. A token can be labeled automatically if it has been labeled before by the user in WebAnno \citep{Yimam14WebAnno} and TALEN \citep{Mayhew18Talen}. In TALEN, a bilingual lexicon can be integrated but just to support annotators that do not speak the text's language. WebAnno and brat \citep{Stenetorp12Brat} allow importing the annotations of external tools as suggestions for the user. The focus is, however, still on the user manually checking all tokens. Also, the annotator cannot use their insight to directly influence and improve the external tool like in the tuning process of ANEA. 
    
In the area of information extraction, the tools by  \citet{Gupta2014SPIED}, \citet{Li2015VINERy} and \citet{Dalvi2016IKE} allow the user to create rules or patterns, e.g. ``[Material] conducts [Energy]''. They can, however, require a large amount of manual rule creation effort to obtain good coverage for NER. With Snorkel \citep{Ratner2019}, a user can define similar and more general labeling functions. \citet{Oiwa2017EntityPopulation} presented a tool to create entity lists manually. These lists could be imported into ANEA. NER is closely related to entity linking. \citet{Zhang18ELISA} presented a system to link entities in many languages automatically but focus on disaster monitoring and, therefore, only consider persons, geopolitical entities, organizations, and locations.

\section{Workflow}
    
The workflow is visualized in Figure \ref{fig:workflow} and we provide an online video that shows an exemplary walkthrough.\footnote{\url{https://www.youtube.com/watch?v=eXwho2Pq6Eg}} The process is split into four parts:

    \begin{figure} \centering
        \begin{subfigure}[b]{0.55\textwidth}
            \includegraphics[width=0.9\textwidth]{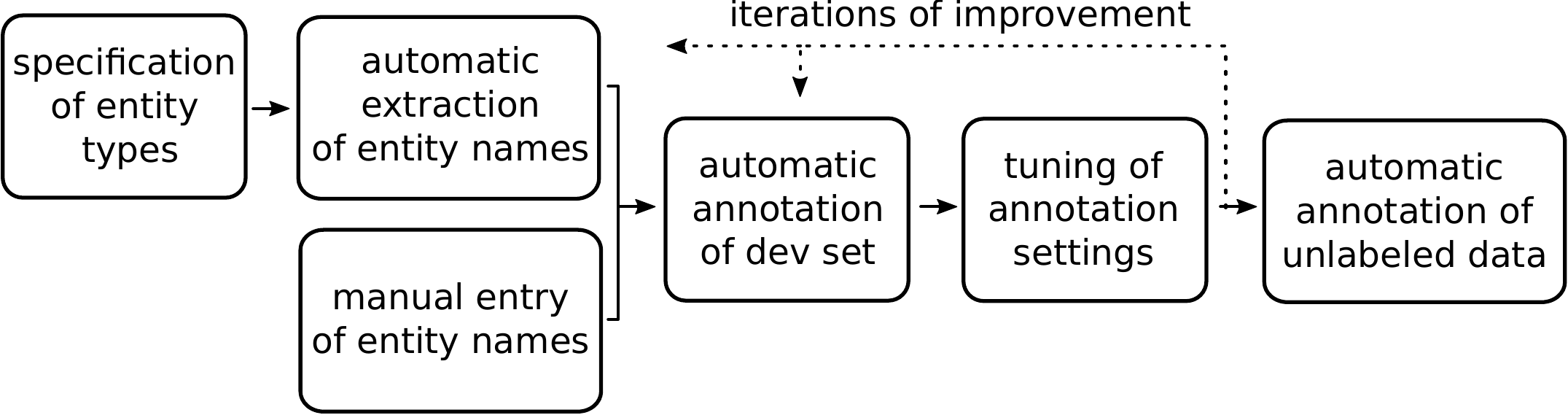}
            \vspace{0.5cm}
            \subcaption{\label{fig:workflow}}
        \end{subfigure}
        \begin{subfigure}[b]{0.35\textwidth}
            \includegraphics[width=\linewidth]{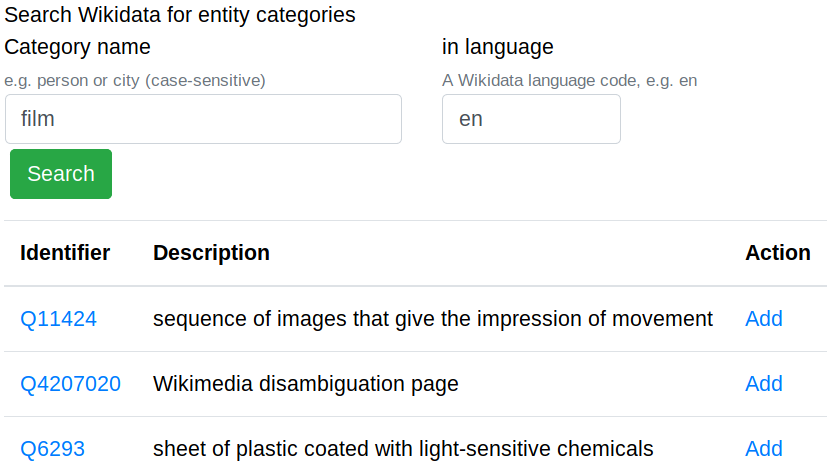}
            \subcaption{\label{fig:screenshot_search}}
        \end{subfigure}
        \vspace{-0.2cm}
        \caption{(a) Overall workflow of ANEA. (b) Interface to search for Wikidata categories from which to extract entity names.}
        \vspace{-0.5cm}
    \end{figure}

\textbf{Extraction}: The user starts by searching for the category names of the entity types that should be extracted (e.g. \textit{person} or \textit{film}). The tool will then automatically extract the names of all the corresponding entities (e.g. for \textit{person}: ``Alan Turing'', ``Edward Sapir'', ...). As the source for the extractions, we use a dump of Wikidata. It is a free and open knowledge base that is created both by manual edits and automatic processes. At the time of writing, it contains over 90 million items. For most items, the names are available in multiple languages (e.g. 32k person names for Yorùbá or 26k movie names for Spanish). The user searches for and specifies the entity types they want to extract and which language should be used for the names (Figure \ref{fig:screenshot_search}). The tool will then extract all items that have the ``is an instance of'' property of the given entity types. The results are the lists of entity names. Additionally, the user can also provide existing lists of entity names in case of a very specific domain.

\textbf{Automatic Annotation}: The automatic annotation is performed by checking each word against the list of extracted entities. A word (or token) is assigned the label of the entity name it matches. If matches of several entity names overlap, the longest match is used. I.e. for the string  ``United Arab Emirates'' the entity name of the country is preferred over the substring ``United'' (the airline) if both are in lists of entities.

\begin{figure} \centering
        \begin{subfigure}[b]{0.4\textwidth}
        \includegraphics[width=\linewidth]{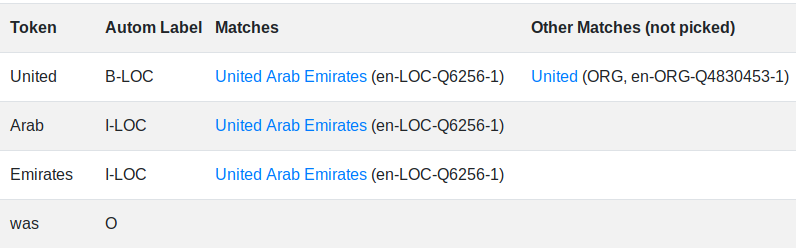}
        \vspace{1.5cm}
        \subcaption{\label{fig:screenshot_inspection}}
        \end{subfigure}
        \begin{subfigure}[b]{0.55\textwidth}
            \centering
            \includegraphics[width=0.9\textwidth]{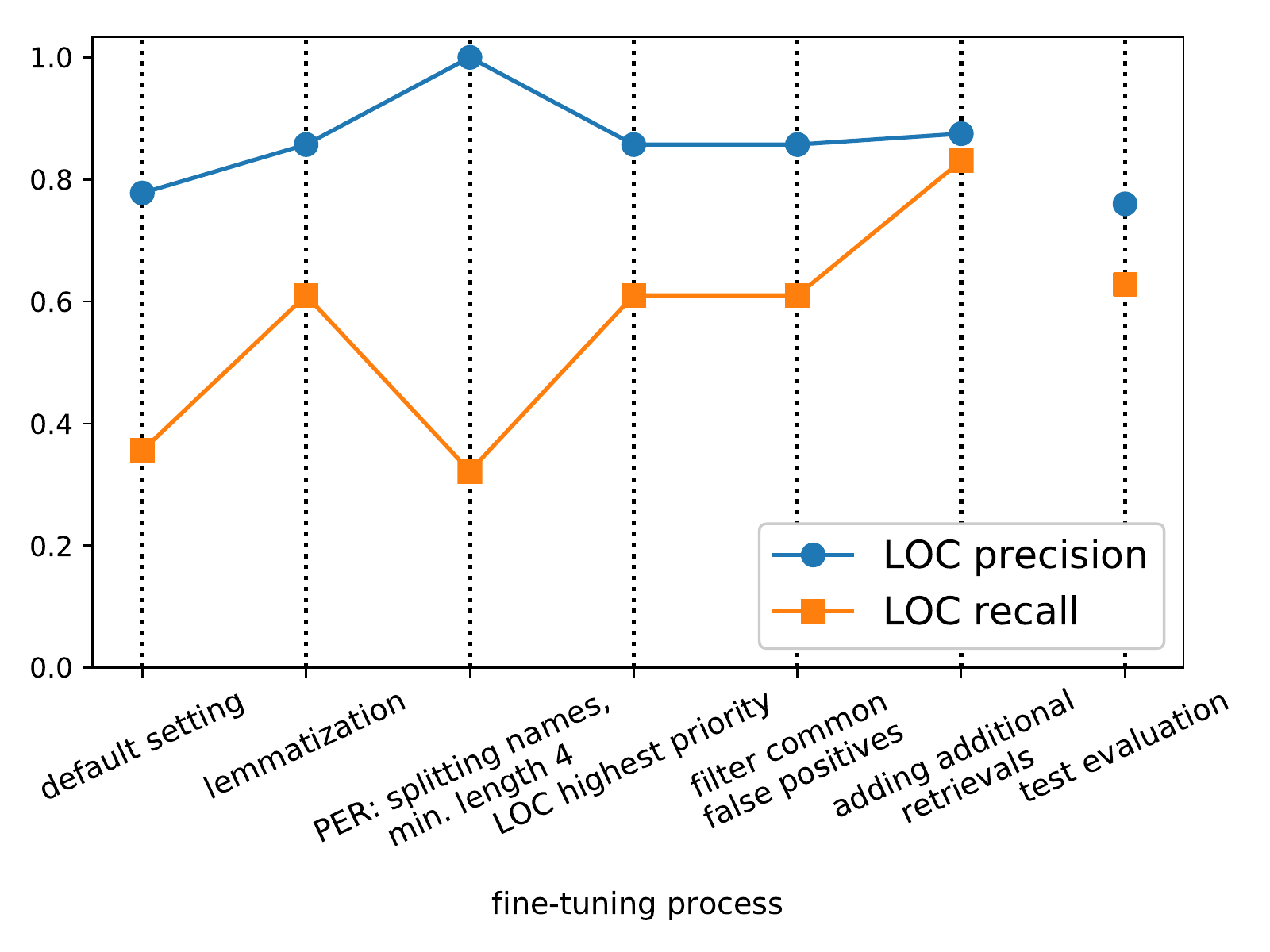}
            \subcaption{\label{fig:tuning}}
        \end{subfigure}
        \caption{(a) Interface to manually inspect the automatic labeling. (b) Development of precision and recall during the tuning process on the Estonian data. On the x-axis, the setting changes over time are reported.}
    \end{figure}

\textbf{Evaluation}: If a small set of labeled data exists, it can be used to evaluate the automatic annotation. The tool can calculate precision, recall and F1-score directly. It also reports the tokens that were most often labeled incorrectly or not labeled. For a more in-depth analysis, for each token, one can check which label was assigned, which alternative labels could have been assigned and to which entities they correspond. This allows a user to easily understand issues of the automatic annotation (Figure \ref{fig:screenshot_inspection}). Specific labels can also be changed manually.

\textbf{Tuning}: ANEA provides multiple options with which the automatic annotation can be improved. Guided by the evaluation from the previous step, this allows the user to easily insert language expertise into the annotation process and prevent common mistakes while still avoiding to annotate or post-edit many tokens manually.  The options include lemmatization, filtering common false positives, stopword removal, adding alias names (like "ICLR" for the "International Conference on Learning Representations"), splitting entity names, removing diacritics, requiring a minimum length for the entities, prioritization of lists for resolution of conflicts or fuzzy matching of entities. 

The effects of such a tuning process are visualized in Figure \ref{fig:tuning} for an Estonian dataset and the \textit{location} label. Adding lemmatization in tuning-step 1 increases recall due to the language's rich morphological structure that can hinder the matching. In step 3, location entities are given a higher priority if they conflict with person entities on the same token. In the last tuning-step, another gain can be obtained by extracting additional entity lists for Estonian locations based on the evaluation feedback. After the (optional) tuning process, unlabeled text can be automatically annotated for use as distant supervision.

\section{Experimental Evaluation}
    \label{sec:experiments}
    
    \subsection{Datasets}
    \label{sec:data}
    
    We selected a variety of datasets that reflect different languages and entity granularities. The first 1500 tokens of each dataset are used as labeled training instances. \citet{garrette-baldridge-2013-learning} reported this as the number of tokens that can be annotated within two hours for a low-resource POS task. We think that this is a reasonable amount of labeled data that one can expect even in a low-resource setting, and it is also necessary for training the baselines we compare to. For \textbf{English (En)}, the CoNLL03 dataset is probably the most popular NER dataset. It was created for the CoNLL-2003 shared task \citep{tjong2003conll}. To obtain a more specialized domain, we manually annotated the \textit{location} labels from the CoNLL03 dataset with more specific labels. For \textbf{Spanish (Es)}, we manually annotated news articles with the label \textit{movie} to resemble a Latin-American setting where e.g. a start-up requires a fine-grained and less common label. For \textbf{Yorùbá (Yo)}, a language spoken predominantly in West Africa, we evaluate on the dataset by \citet{alabi2019massive}. We also evaluate on two European low-resource languages, namely \textbf{Estonian (Et)} \citep{Tkachenko13Estonian} and \textbf{West Frisian (Fy)} \citep{Pan17CrossLingual}. All results are reported on held-out test sets. The manually labeled data created for this evaluation will be made publicly available.
    
    \subsection{Machine Learning Models}
    
    We evaluate against two baselines that should, like ANEA, be easy and quick to use, do not require extensive development of hand-engineered features and do not have large hardware requirements. The Stanford NER tagger \citep{Finkel05StanfordTagger} is a popular tool based on Conditional-Random-Fields (\textbf{CRF}) which we use in their suggested configuration\footnote{\url{https://nlp.stanford.edu/software/crf-faq.html\#a}}. For the second baseline, a neural network (\textbf{NN}), we performed preliminary experiments on held-out, English data in a low-resource setting and chose a combination of a bidirectional Gated Recurrent Unit \citep{Cho14GRU} and a ReLU with Dropout \citep{Srivastava14Dropout} between the layers. To easily apply the model to many different languages, we used pretrained fastText embeddings \citep{grave2018learning} which are available in 157 languages. Model details are given in the code. In the high-resource setting on the full CoNLL03 dataset ($>$250k labeled tokens), both baselines achieve an F1-score of 87.
    
    \subsection{Experimental Setup}
    \textbf{Experiment A:} Here, the quality of the automatic annotation is evaluated. The CRF is trained on the 1500 labeled training tokens of each dataset. Similarly, for the neural network, the first 1000 tokens are used for the training. The remaining 500 tokens are held-out as the development set to select the best performing epoch and avoid overfitting. For ANEA, we report the scores with and without the tuning phase. \textit{ANEA No Tuning} just uses the default settings without any labeled supervision and no manual interaction. For \textit{ANEA + Tuning}, the 1500 labeled training token are used for the manual tuning. The manual steps were performed by a subject with experience in NLP and fluency in English and Spanish but not in the other languages. Interaction was limited to no more than 10 manual steps and 30 minutes of user interaction per dataset. 
    
    \textbf{Experiment B:} For evaluating the effect of the distant supervision, unlabeled tokens are automatically annotated by the CRF, the NN and ANEA with Tuning. The NN model is then retrained on both the manually labeled and the distantly-supervised instances. 200k tokens from each of the datasets are used as unlabeled data. For Spanish, West Frisian and Yorùbá, ca. 15k and 70k and 18k tokens are used, respectively, due to the smaller dataset sizes. These texts are disjoint of the labeled training and test data.
    
    \subsubsection{Results}
    The results of Experiment A are given in Table~\ref{tab:results}. The CRF approach can provide a high precision but often has a very low recall due to the limited amount of training data. The NN can leverage the pre-training of the embeddings on large amounts of unlabeled text. However, the training data seems not enough to reach a competitive performance. Our tool struggles most with organizations as these are stored as several different entity types in Wikidata. Another issue is the existence of false positives of words that have other meanings beyond entity names, e.g. the Turkish city ``Of''. Nevertheless, reasonable results are obtained even if the amount of labeled tokens is too low for the baselines to learn anything meaningful (cf. \textit{En CONTINENT} or \textit{Et ORG}). Even without any labeled data, we are often able to reach competitive performance. Using the tuning process is helpful to boost the performance further. The possibility for the user to trade-off precision and recall can be seen in several cases (e.g. \textit{En LOC} or \textit{Et PER}).  Overall, ANEA outperforms the other baselines in all metrics in a majority of the settings. It achieves the best F1-score in all but one case.
    
    The higher quality of the automatic annotation is also reflected in Experiment B. For 14 out of 16 evaluated entity types, the distant supervision provided by ANEA achieves the largest improvements. On average, it increases the classifier's performance by 18 points F1-score.
    
    \begin{table}
        \begin{subtable}{0.6\textwidth}
            \scriptsize
            \centering
            \setlength\tabcolsep{3pt}
            \begin{tabular}{l c c c c}
                \toprule
                & CRF & NN & ANEA & ANEA \\
                &     &        & No Tuning  & + Tuning \\ \midrule
                &   $\,$P$\;\;\,$R$\;\,$F1 &  $\,$P$\;\;\,$R$\;\,$F1 & $\,$P$\;\;\,$R$\;\,$F1 & $\,$P$\;\;\,$R$\;\,$F1  \\  
                En PER & \textbf{75} 14 23 & 54 40 46  & 36 \textbf{51} 42 & 67 49 \textbf{57} \\ 
                En LOC & 66 22 33 & 54 52 52 & \textbf{70} 45 55 & 56 \textbf{74} \textbf{64} \\
                En ORG & \textbf{24} 08 12 & 23 \textbf{13} \textbf{16} & 17 07 10 & 21 09 13 \\ 
                \midrule
                En CITY & \textbf{100} 14 25 & 27 43 33 & 16 30 21 & 29 \textbf{51} \textbf{37} \\ 
                En COUN. & \textbf{94} 05 10 & 63 51 56 & 93 80 86 & 84 \textbf{90} \textbf{87} \\
                En CONTI. & 00 00 00 & 00 00 00 & \textbf{75} \textbf{94} \textbf{83} & \textbf{75} \textbf{94} \textbf{83} \\ \midrule
                Es MOVIE & \textbf{75} 02 05 & 08 07 08 & 32 35 33 & 40 \textbf{40} \textbf{40} \\ \midrule  
                Et PER & 66 24 35 & 61 30 40 & \textbf{75} 17 27 & 41 \textbf{51} \textbf{45} \\  
                Et LOC & 59 27 37 & 44 25 32 & 71 36 48 & \textbf{76} \textbf{63} \textbf{69} \\
                Et ORG & 00 00 00 & 17 09 12 & 75 12 21 & \textbf{81} \textbf{17} \textbf{29} \\
                \midrule
                Fy PER & 07 06 07 & 04 03 04 & \textbf{55} \textbf{42} \textbf{48} & \textbf{55} \textbf{42} \textbf{48}\\ 
                Fy LOC & 32 55 41 & 33 42 37 & \textbf{68} 24 37 & 61 \textbf{34} \textbf{43}\\
                Fy ORG & 00 00 00 & 00 00 00 & 89 07 13 & \textbf{90} \textbf{08} \textbf{14}\\ \bottomrule
                Yo PER & 33 05 10 & 15 22 18 & 11 13 12 & \textbf{49} \textbf{43} \textbf{46} \\
                Yo LOC & \textbf{100} 07 12 & 48 27 35 & 64 72 68 & 65 \textbf{74} \textbf{69} \\
                Yo ORG & 00 00 00 & 07 08 08 & 16 28 20 & \textbf{46} \textbf{52} \textbf{49} \\
              \bottomrule
            \end{tabular}\\ \vspace{0.2cm}
           { \small Experiment A }
            \end{subtable}
            \begin{subtable}{0.3\textwidth}
                \scriptsize
                \centering
                \setlength\tabcolsep{3pt}
                \begin{tabular}{l m{0.7cm} m{0.7cm} m{0.7cm}}
                    \toprule
                    \multicolumn{4}{c}{NN  + Distant Supervision by ...} \\ 
                    &   CRF & NN & ANEA  \\  \midrule
                    En PER & -35 & +5 & \textbf{+15} \\
                    En LOC & -20 & +1 & \textbf{+13} \\
                    En ORG & -6 & \textbf{0} & -5 \\
                    \midrule
                    En CITY & -13 & +1 & \textbf{+6} \\
                    En COUN. & -45 & -6 & \textbf{+30} \\
                    En CONTI. & 0 & 0 & \textbf{+88} \\
                    \midrule
                    Es Movie & -7 & +2 & \textbf{+14} \\
                    \midrule 
                    Et PER & -7 & -7 & \textbf{+14} \\
                    Et LOC & +10 & -1 & \textbf{+39} \\
                    Et ORG & -2 & 0 & \textbf{+17} \\
                    \midrule
                    Fy PER & +1 & 0 & \textbf{+26} \\
                    Fy LOC & \textbf{+4} & +1 & \textbf{+4} \\
                    Fy ORG & +1 & +1 & \textbf{+7} \\
                    \midrule
                    Yo PER & -4 & \textbf{+6} & -5 \\
                    Yo LOC & -25 & +4 & \textbf{+5} \\
                    Yo ORG & -1 & +1 & \textbf{+20} \\ \bottomrule
                \end{tabular}
                 \\ \vspace{0.2cm}
           { \small Experiment B } 
        \end{subtable}
        \vspace{-0.25cm}
        \caption{Results of Experiment A and Experiment B on the test data. We report precision/recall/F1-score in percentage (higher is better).\label{tab:results}}
        \vspace{-0.4cm}
    \end{table}
    
    \section{Technical Aspects}
    \vspace{-0.2cm}
    
    The tool consists of both a library for the core functionalities as well as a graphical user interface. The user  can control the interface in the browser with the back end running on the local system. Alternatively, the back end can run on a different, more powerful machine and is then accessed remotely. All the code is published as open-source under the Apache 2 license, and we welcome contributions from other authors. The tool is implemented in Python 3 using Flask\footnote{\url{http://flask.pocoo.org}} for the webserver's back end and Bootstrap 4\footnote{\url{https://getbootstrap.com}} for the front end. To overcome the rate limitations of the Wikidata Web API, a database dump of Wikidata is used. To reduce hardware requirements, care was taken during the implementation to limit the memory footprint.
    
    The user can upload text files or insert them directly into a text field. For labeled data, the CoNLL column format is supported. Annotated text can be downloaded in the same format. Tokenization and lemmatization are provided for a variety of languages via the SpaCy \citep{spacy} and EstNLTK \citep{laur-EtAl:2020:LREC}. For other languages, the text can be preprocessed with an external system before inputting it, or the external tool can be easily integrated into ANEA. Stopword lists for 58 languages are included.

    \section{Conclusion}
    \vspace{-0.2cm}
    
    We presented an open-source tool to obtain large amounts of distantly supervised training data for NER in a quick way and with few manual efforts and costs. While the annotation itself is automatic, the user can tune it to add their expertise. To support users of varying technical backgrounds, both a library and a graphical user interface are provided. The experiments showed its usefulness in six different language and domain settings. 
    
    ANEA has already been used successfully to provide the distant supervision in the works by \cite{lange-etal-2019-nlnde}, \cite{Lange2019FeatureDependent},  \cite{adelani2020distant}, \cite{hedderich-etal-2020-transfer} and \cite{hedderich2021noisemodelerror}. In the future, we aim to add techniques from active learning to improve the  leverage of expert insights further. Previous work has also shown that performance gains through distant supervision can be boosted by handling errors in the automatic annotation via label noise handling. We see the integration of these approaches as an additional avenue for interesting future work. 

\section*{Acknowledgments}
This work has been partially funded by the Deutsche Forschungsgemeinschaft (DFG, German Research Foundation) – Project-ID 232722074 – SFB 1102 and the EU Horizon 2020 project ROXANNE under grant number 833635.
\bibliography{anea}

\begin{thebibliography}{31}
\providecommand{\natexlab}[1]{#1}
\providecommand{\url}[1]{\texttt{#1}}
\expandafter\ifx\csname urlstyle\endcsname\relax
  \providecommand{\doi}[1]{doi: #1}\else
  \providecommand{\doi}{doi: \begingroup \urlstyle{rm}\Url}\fi

\bibitem[Adelani et~al.(2020)Adelani, Hedderich, Zhu, van~den Berg, and
  Klakow]{adelani2020distant}
David~Ifeoluwa Adelani, Michael~A. Hedderich, Dawei Zhu, Esther van~den Berg,
  and Dietrich Klakow.
\newblock Distant supervision and noisy label learning for low resource named
  entity recognition: {A} study on hausa and yor{\`{u}}b{\'{a}}.
\newblock \emph{CoRR}, abs/2003.08370, 2020.
\newblock URL \url{https://arxiv.org/abs/2003.08370}.

\bibitem[Alabi et~al.(2020)Alabi, Amponsah-Kaakyire, Adelani, and
  Espa{\~n}a-Bonet]{alabi2019massive}
Jesujoba~O Alabi, Kwabena Amponsah-Kaakyire, David~I Adelani, and Cristina
  Espa{\~n}a-Bonet.
\newblock Massive vs. curated word embeddings for low-resourced languages. the
  case of yorùbá and {Twi}.
\newblock In \emph{Proc. of LREC 2020}, 2020.

\bibitem[Cho et~al.(2014)Cho, van Merrienboer, Gulcehre, Bahdanau, Bougares,
  Schwenk, and Bengio]{Cho14GRU}
Kyunghyun Cho, Bart van Merrienboer, Caglar Gulcehre, Dzmitry Bahdanau, Fethi
  Bougares, Holger Schwenk, and Yoshua Bengio.
\newblock Learning phrase representations using rnn encoder--decoder for
  statistical machine translation.
\newblock In \emph{Proc. of EMNLP 2014}, 2014.
\newblock URL \url{http://aclweb.org/anthology/D14-1179}.

\bibitem[Dalvi et~al.(2016)Dalvi, Bhakthavatsalam, Clark, Clark, Etzioni,
  Fader, and Groeneveld]{Dalvi2016IKE}
Bhavana Dalvi, Sumithra Bhakthavatsalam, Chris Clark, Peter Clark, Oren
  Etzioni, Anthony Fader, and Dirk Groeneveld.
\newblock {IKE} - an interactive tool for knowledge extraction.
\newblock In \emph{Proc. of AKBC 2016}, 2016.
\newblock \doi{10.18653/v1/W16-1303}.

\bibitem[Finkel et~al.(2005)Finkel, Grenager, and
  Manning]{Finkel05StanfordTagger}
Jenny~Rose Finkel, Trond Grenager, and Christopher Manning.
\newblock Incorporating non-local information into information extraction
  systems by gibbs sampling.
\newblock In \emph{Proc. of ACL 2005}, 2005.
\newblock URL \url{http://aclweb.org/anthology/P05-1045}.

\bibitem[Garrette \& Baldridge(2013)Garrette and
  Baldridge]{garrette-baldridge-2013-learning}
Dan Garrette and Jason Baldridge.
\newblock Learning a part-of-speech tagger from two hours of annotation.
\newblock In \emph{Proc. of NAACL 2013}, 2013.
\newblock URL \url{https://www.aclweb.org/anthology/N13-1014}.

\bibitem[Grave et~al.(2018)Grave, Bojanowski, Gupta, Joulin, and
  Mikolov]{grave2018learning}
Edouard Grave, Piotr Bojanowski, Prakhar Gupta, Armand Joulin, and Tomas
  Mikolov.
\newblock Learning word vectors for 157 languages.
\newblock In \emph{Proc. of LREC 2018}, 2018.
\newblock URL \url{https://www.aclweb.org/anthology/L18-1550}.

\bibitem[Gupta \& Manning(2014)Gupta and Manning]{Gupta2014SPIED}
Sonal Gupta and Christopher Manning.
\newblock {SPIED}: {S}tanford pattern based information extraction and
  diagnostics.
\newblock In \emph{Proc. of the Workshop on Interactive Language Learning,
  Visualization, and Interfaces}, 2014.
\newblock \doi{10.3115/v1/W14-3106}.

\bibitem[Hedderich et~al.(2020)Hedderich, Adelani, Zhu, Alabi, Markus, and
  Klakow]{hedderich-etal-2020-transfer}
Michael~A. Hedderich, David Adelani, Dawei Zhu, Jesujoba Alabi, Udia Markus,
  and Dietrich Klakow.
\newblock Transfer learning and distant supervision for multilingual
  transformer models: A study on {A}frican languages.
\newblock In \emph{Proc. of EMNLP 2020}, 2020.
\newblock URL \url{https://www.aclweb.org/anthology/2020.emnlp-main.204}.

\bibitem[Hedderich et~al.(2021{\natexlab{a}})Hedderich, Lange, Adel, Strötgen,
  and Klakow]{hedderich2020survey}
Michael~A. Hedderich, Lukas Lange, Heike Adel, Jannik Strötgen, and Dietrich
  Klakow.
\newblock A survey on recent approaches for natural language processing in
  low-resource scenarios.
\newblock In \emph{Proc. of NAACL 2021}, 2021{\natexlab{a}}.
\newblock URL \url{https://arxiv.org/abs/2010.12309}.

\bibitem[Hedderich et~al.(2021{\natexlab{b}})Hedderich, Zhu, and
  Klakow]{hedderich2021noisemodelerror}
Michael~A. Hedderich, Dawei Zhu, and Dietrich Klakow.
\newblock Analysing the noise model error for realistic noisy label data.
\newblock \emph{CoRR}, abs/2101.09763, 2021{\natexlab{b}}.
\newblock URL \url{https://arxiv.org/abs/2101.09763}.

\bibitem[Honnibal et~al.(2020)Honnibal, Montani, Van~Landeghem, and
  Boyd]{spacy}
Matthew Honnibal, Ines Montani, Sofie Van~Landeghem, and Adriane Boyd.
\newblock {spaCy: Industrial-strength Natural Language Processing in Python},
  2020.
\newblock URL \url{https://doi.org/10.5281/zenodo.1212303}.

\bibitem[Kann et~al.(2020)Kann, Lacroix, and S{\o}gaard]{kann2020POSPoorly}
Katharina Kann, Oph{\'{e}}lie Lacroix, and Anders S{\o}gaard.
\newblock Weakly supervised {POS} taggers perform poorly on \emph{Truly}
  low-resource languages.
\newblock In \emph{Proc. of {AAAI} 2020}, 2020.
\newblock URL \url{https://aaai.org/ojs/index.php/AAAI/article/view/6317}.

\bibitem[Lange et~al.(2019{\natexlab{a}})Lange, Adel, and
  Str{\"o}tgen]{lange-etal-2019-nlnde}
Lukas Lange, Heike Adel, and Jannik Str{\"o}tgen.
\newblock {NLNDE}: Enhancing neural sequence taggers with attention and noisy
  channel for robust pharmacological entity detection.
\newblock In \emph{Proc. of The 5th Workshop on BioNLP Open Shared Tasks},
  2019{\natexlab{a}}.
\newblock URL \url{https://www.aclweb.org/anthology/D19-5705}.

\bibitem[Lange et~al.(2019{\natexlab{b}})Lange, Hedderich, and
  Klakow]{Lange2019FeatureDependent}
Lukas Lange, Michael~A. Hedderich, and Dietrich Klakow.
\newblock Feature-dependent confusion matrices for low-resource ner labeling
  with noisy labels.
\newblock In \emph{Proc. of EMNLP 2019}, 2019{\natexlab{b}}.

\bibitem[Laur et~al.(2020)Laur, Orasmaa, Särg, and Tammo]{laur-EtAl:2020:LREC}
Sven Laur, Siim Orasmaa, Dage Särg, and Paul Tammo.
\newblock Estnltk 1.6: Remastered estonian nlp pipeline.
\newblock In \emph{Proc. of LREC 2020}, 2020.
\newblock URL \url{https://www.aclweb.org/anthology/2020.lrec-1.884}.

\bibitem[Li et~al.(2015)Li, Kim, Touchette, Venkatachalam, and
  Wang]{Li2015VINERy}
Yunyao Li, Elmer Kim, Marc~A. Touchette, Ramiya Venkatachalam, and Hao Wang.
\newblock Vinery: A visual ide for information extraction.
\newblock \emph{Proc. of the VLDB Endowment}, 8\penalty0 (12), 2015.
\newblock \doi{10.14778/2824032.2824108}.

\bibitem[Lison et~al.(2020)Lison, Barnes, Hubin, and
  Touileb]{lison-etal-2020-weak-supervision}
Pierre Lison, Jeremy Barnes, Aliaksandr Hubin, and Samia Touileb.
\newblock Named entity recognition without labelled data: A weak supervision
  approach.
\newblock In \emph{Proc of ACL 2020}, 2020.
\newblock URL \url{https://www.aclweb.org/anthology/2020.acl-main.139}.

\bibitem[Mayhew \& Roth(2018)Mayhew and Roth]{Mayhew18Talen}
Stephen Mayhew and Dan Roth.
\newblock Talen: Tool for annotation of low-resource entities.
\newblock In \emph{Proc. of ACL 2018: System Demonstrations}, 2018.
\newblock URL \url{http://aclweb.org/anthology/P18-4014}.

\bibitem[Mintz et~al.(2009)Mintz, Bills, Snow, and Jurafsky]{mintz2009distant}
Mike Mintz, Steven Bills, Rion Snow, and Daniel Jurafsky.
\newblock Distant supervision for relation extraction without labeled data.
\newblock In \emph{Proc. of ACL}, 2009.
\newblock URL \url{https://www.aclweb.org/anthology/P09-1113}.

\bibitem[Oiwa et~al.(2017)Oiwa, Suhara, Komiya, and
  Lopatenko]{Oiwa2017EntityPopulation}
Hidekazu Oiwa, Yoshihiko Suhara, Jiyu Komiya, and Andrei Lopatenko.
\newblock A lightweight front-end tool for interactive entity population.
\newblock \emph{CoRR}, abs/1708.00481, 2017.
\newblock URL \url{http://arxiv.org/abs/1708.00481}.

\bibitem[Pan et~al.(2017)Pan, Zhang, May, Nothman, Knight, and
  Ji]{Pan17CrossLingual}
Xiaoman Pan, Boliang Zhang, Jonathan May, Joel Nothman, Kevin Knight, and Heng
  Ji.
\newblock Cross-lingual name tagging and linking for 282 languages.
\newblock In \emph{Proc. of ACL 2017}, 2017.
\newblock \doi{10.18653/v1/P17-1178}.

\bibitem[Peng et~al.(2019)Peng, Xing, Zhang, Fu, and
  Huang]{peng-etal-2019-distantly}
Minlong Peng, Xiaoyu Xing, Qi~Zhang, Jinlan Fu, and Xuanjing Huang.
\newblock Distantly supervised named entity recognition using
  positive-unlabeled learning.
\newblock In \emph{Proc. of ACL 2019}, 2019.
\newblock \doi{10.18653/v1/P19-1231}.
\newblock URL \url{https://www.aclweb.org/anthology/P19-1231}.

\bibitem[Ratner et~al.(2019)Ratner, Bach, Ehrenberg, Fries, Wu, and
  R{\'e}]{Ratner2019}
Alexander Ratner, Stephen~H. Bach, Henry Ehrenberg, Jason Fries, Sen Wu, and
  Christopher R{\'e}.
\newblock Snorkel: rapid training data creation with weak supervision.
\newblock \emph{The VLDB Journal}, Jul 2019.
\newblock ISSN 0949-877X.
\newblock \doi{10.1007/s00778-019-00552-1}.

\bibitem[Srivastava et~al.(2014)Srivastava, Hinton, Krizhevsky, Sutskever, and
  Salakhutdinov]{Srivastava14Dropout}
Nitish Srivastava, Geoffrey Hinton, Alex Krizhevsky, Ilya Sutskever, and Ruslan
  Salakhutdinov.
\newblock Dropout: A simple way to prevent neural networks from overfitting.
\newblock \emph{J. Mach. Learn. Res.}, 15\penalty0 (1), 2014.

\bibitem[Stenetorp et~al.(2012)Stenetorp, Pyysalo, Topi{\'{c}}, Ohta,
  Ananiadou, and Tsujii]{Stenetorp12Brat}
Pontus Stenetorp, Sampo Pyysalo, Goran Topi{\'{c}}, Tomoko Ohta, Sophia
  Ananiadou, and Jun'ichi Tsujii.
\newblock brat: a web-based tool for nlp-assisted text annotation.
\newblock In \emph{Proc. of the Demonstrations at EACL 2012}, 2012.
\newblock URL \url{http://aclweb.org/anthology/E12-2021}.

\bibitem[Tjong Kim~Sang \& De~Meulder(2003)Tjong Kim~Sang and
  De~Meulder]{tjong2003conll}
Erik~F Tjong Kim~Sang and Fien De~Meulder.
\newblock Introduction to the conll-2003 shared task: Language-independent
  named entity recognition.
\newblock In \emph{Proc. of the Seventh Conference on Natural Language
  Learning}, 2003.
\newblock URL \url{https://www.aclweb.org/anthology/W03-0419}.

\bibitem[Tkachenko et~al.(2013)Tkachenko, Petmanson, and
  Laur]{Tkachenko13Estonian}
Alexander Tkachenko, Timo Petmanson, and Sven Laur.
\newblock Named entity recognition in estonian.
\newblock In \emph{Proc. of the 4th Biennial International Workshop on
  Balto-Slavic Natural Language Processing}, 2013.

\bibitem[Weischedel et~al.(2011)Weischedel, Pradhan, Ramshaw, Palmer, Xue,
  Marcus, Taylor, Greenberg, Hovy, Belvin, et~al.]{weischedel2011ontonotes}
Ralph Weischedel, Sameer Pradhan, Lance Ramshaw, Martha Palmer, Nianwen Xue,
  Mitchell Marcus, Ann Taylor, Craig Greenberg, Eduard Hovy, Robert Belvin,
  et~al.
\newblock Ontonotes release 4.0.
\newblock \emph{LDC2011T03}, 2011.

\bibitem[Yimam et~al.(2014)Yimam, Biemann, Eckart~de Castilho, and
  Gurevych]{Yimam14WebAnno}
Seid~Muhie Yimam, Chris Biemann, Richard Eckart~de Castilho, and Iryna
  Gurevych.
\newblock Automatic annotation suggestions and custom annotation layers in
  webanno.
\newblock In \emph{Proc. of ACL 2014: System Demonstrations}, 2014.

\bibitem[Zhang et~al.(2018)Zhang, Lin, Pan, Lu, May, Knight, and
  Ji]{Zhang18ELISA}
Boliang Zhang, Ying Lin, Xiaoman Pan, Di~Lu, Jonathan May, Kevin Knight, and
  Heng Ji.
\newblock Elisa-edl: A cross-lingual entity extraction, linking and
  localization system.
\newblock In \emph{Proc. of NAACL-HLT 2018: Demonstrations}, 2018.
\newblock URL \url{http://aclweb.org/anthology/N18-5009}.

\end{thebibliography}
\bibliographystyle{iclr2020_conference}

\end{document}